\documentclass[10pt,twocolumn,letterpaper]{article}
\usepackage[pagenumbers]{cvpr}
\usepackage{graphicx}
\usepackage{amsmath}
\usepackage{amssymb}
\usepackage{booktabs}
\usepackage{dsfont} % for /mathds
\usepackage{pifont}% http://ctan.org/pkg/pifont xmark and cmark
\newcommand{\cmark}{\ding{51}}%
\usepackage[pagebackref,breaklinks,colorlinks]{hyperref}

\usepackage[capitalize]{cleveref}
\crefname{section}{Sec.}{Secs.}
\Crefname{section}{Section}{Sections}
\Crefname{table}{Table}{Tables}
\crefname{table}{Tab.}{Tabs.}

\begin{document}

%%%%%%%%% TITLE - PLEASE UPDATE
\title{Towards Panoptic 3D Parsing for Single Image in the Wild}

\author{Sainan Liu$^{1,2}$ \quad\quad Vincent Nguyen$^1$ \quad\quad Yuan Gao$^1$ \quad\quad Subarna Tripathi$^2$ \quad\quad Zhuowen Tu$^1$\\
$^1$UC San Diego \quad\quad $^2$Intel Labs\\
{\tt\small \{sal131, vvn012, y1gao, ztu\}@ucsd.edu, \{sainan.liu, subarna.tripathi\}@intel.com}
}
\maketitle

%%%%%%%%% ABSTRACT
\begin{abstract}
Performing single image holistic understanding and 3D reconstruction is a central task in computer vision. This paper presents an integrated system that performs dense scene labeling, object detection, instance segmentation, depth estimation, 3D shape reconstruction, and 3D layout estimation for indoor and outdoor scenes from a single RGB image. We name our system panoptic 3D parsing (Panoptic3D) in which panoptic segmentation (``stuff'' segmentation and ``things'' detection/segmentation) with 3D reconstruction is performed. We design a stage-wise system, Panoptic3D (stage-wise), where a complete set of annotations is absent. Additionally, we present an end-to-end pipeline, Panoptic3D (end-to-end), trained on a synthetic dataset with a full set of annotations. We show results on both indoor (3D-FRONT) and outdoor (COCO and Cityscapes) scenes. Our proposed panoptic 3D parsing framework points to a promising direction in computer vision. Panoptic3D can be applied to a variety of applications, including autonomous driving, mapping, robotics, design, computer graphics, robotics, human-computer interaction, and augmented reality.
\end{abstract}
\begin{figure}[!ht]
    \centering
    \includegraphics[width=\linewidth]{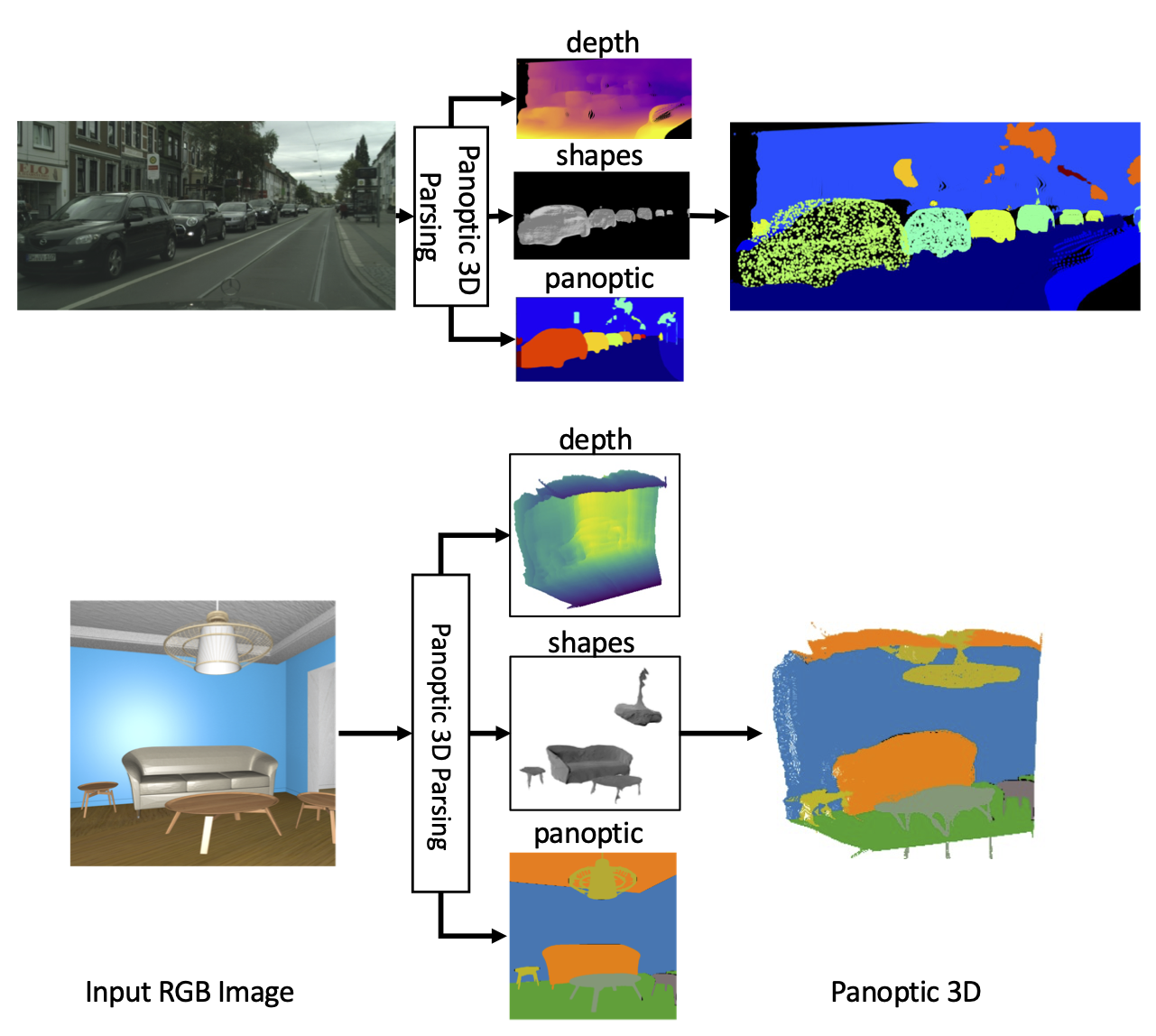}
    \caption{Our stage-wise Panoptic 3D Parsing System (Panoptic3D), shown in the first row, takes a single-view image as input and outputs panoptic 3D results, a scene reconstruction with ``stuff'' and ``things'' shown on the right-hand side. The stage-wise pipeline works on natural image datasets where complete ground-truth annotations for segmentation and 3D reconstruction are absent, such as COCO \cite{Lin2014COCO} and Cityscapes \cite{Cordts2016Cityscapes} where ground-truth 3D reconstruction is missing in both training and evaluation. The second row shows an alternative end-to-end model, Panoptic3D (end-to-end) applied to the synthetic 3D-FRONT \cite{fu20203dfuture} dataset.}
    \label{fig:teaser}
    \vspace{-3mm}
\end{figure}
%%%%%%%%% BODY TEXT
\section{Introduction}
\label{sec:intro}
One of the main objectives in computer vision is to develop systems that can ``see'' the world \cite{marr1982vision,tarr2002visual}. Here we tackle the problem of single image holistic understanding and 3D reconstruction, which is deeply rooted in decades of development in computer vision and photogrammetry but became practically feasible only recently thanks to the exploding growth in modeling, computing \cite{krizhevsky2012imagenet,szegedy2015going,simonyan2015very,he2016deep,goodfellow2016deep,xie2017aggregated}, and large-scale datasets \cite{deng2009imagenet,Lin2014COCO,Cordts2016Cityscapes,chang2015shapenet,fu20203dfuture}.

\begin{table*}[!ht]
\vspace{-2mm}
\begin{center}
   \caption{\small Comparison for different 3D reconstruction methods.
   Mesh-RCNN \cite{gkioxari2019meshrcnn} only allows single-instance per image during training, but it will enable outputs of multi-object components at inference time. Nevertheless, efforts are still required to allow for the multi-object module in an end-to-end pipeline for training and evaluation.
   }
    \label{tab:method_comparison}
\scalebox{0.90}{
    \begin{tabular}{ccccccc}
    \hline
    Method &3D & Single & Layout & Panoptic & Outdoor &Multiple \\
    & & image & 3D & segmentation & scenes & objects \\
    \hline
    \cite{zhang2018genre,kanazawa2018end,groueix2018atlas,xu2019disn} & \cmark & \cmark & & & &\\
    \cite{gkioxari2019meshrcnn} & \cmark & \cmark & & & & $\dagger$\\
    \cite{han2004automatic} & \cmark & \cmark & & &  \cmark & \\
    \cite{Tulsiani17factored3d,zou2018layoutnet,huang2018cooperative,nie2020total3d} & \cmark & \cmark & \cmark & & & \cmark\\
   \cite{Li2018megadepth} & \cmark & \cmark & & & \cmark &\\
    \cite{Alhashim2018densedepth}&\cmark&\cmark&\cmark& &\cmark&\\
    \cite{pollefeys1999self,pollefeys2008detailed} & \cmark & &\cmark & & \cmark & \cmark\\
    \cite{Kirillov2018panoptic,Kirillov2019fpn,xiong19upsnet,Lazarow2019ocfusion} & & \cmark & & \cmark & \cmark & \cmark\\
    \hline
    {\small Panoptic3DParsing (ours)} & \cmark & \cmark & \cmark & \cmark & \cmark & \cmark\\ 
    \hline
    \end{tabular}
}
\end{center}
\vspace{-3mm}
\end{table*}

We name our system single image panoptic 3D parsing (Panoptic3D). It takes in a single natural RGB image and jointly performs dense image semantic labeling \cite{shotton2006textonboost,tu2008auto,long2015fully},  object detection \cite{Ren2015FasterRCNN},  instance segmentation \cite{he2017mask}, depth estimation \cite{Li2018megadepth, Alhashim2018densedepth}, object shape 3D reconstruction \cite{zhang2018genre, gkioxari2019meshrcnn}, and 3D layout estimation \cite{Tulsiani17factored3d} from a single natural RGB image.
Figure \ref{fig:teaser} gives an illustration for the pipeline where a 3D scene is estimated from a single-view RGB image with the background layout (``stuff'') segmented and the individual foreground instances (``things'') detected, segmented, and fully reconstructed. A closely related work to our Panoptic3D is the recent Total3DUnderstanding method \cite{nie2020total3d} where 3D layout and instance 3D meshes are reconstructed for an indoor input image. However, Total3DUnderstanding \cite{nie2020total3d} does not perform panoptic segmentation \cite{Kirillov2018panoptic}/image parsing \cite{tu2005image} and is limited to indoor scenes.

We briefly discuss the literature from two main angles: 1). 3D reconstruction, particularly from a single-view RGB image; and 2). image understanding, particularly for panoptic segmentation \cite{Kirillov2018panoptic} and image parsing \cite{tu2005image}.

3D reconstruction is an important area in photogrammetry \cite{linder2009digital,colomina2014unmanned} and computer vision \cite{hartley2003multiple,ma2012invitation, ullman1979interpretation,brown2003advances,pollefeys1999self,kutulakos2000theory,pollefeys2008detailed,szeliski2010computer}.
We limit our scope to single RGB image input for 3D instance reconstruction \cite{wu2015shapenets,zhang2018genre,wang2018pixel2mesh,groueix2018atlas,kanazawa2018end,xu2019disn,chen2020topology} and 3D layout generation \cite{Tulsiani17factored3d,zou2018layoutnet}.
There is a renewed interest in performing holistic image and object segmentation (Image Parsing \cite{tu2005image}), called panoptic segmentation \cite{Kirillov2018panoptic,Kirillov2019fpn,xiong19upsnet,Lazarow2019ocfusion}, where the background regions (``stuff'') are labeled with the foreground objects (``things'') detected. Our panoptic 3D parsing method is a system that gives holistic 3D scene reconstruction and understanding for an input image. It includes multiple tasks such as depth estimation, panoptic segmentation, and object instance reconstruction. 

In Section \ref{sec:related}, we discuss the motivations for the individual modules. A comparison between the existing methods and ours is illustrated in Table \ref{tab:method_comparison}. The contributions of our work are summarized below.
\begin{itemize}
 \setlength\itemsep{0mm}
    \item We present a stage-wise system for panoptic 3D parsing, Panoptic3D (stage-wise), by comabatting the issue where full annotations for panoptic segmentation, depth, and 3D instances are absent. To the best of our knowledge, this is the first system of its kind to perform joint panoptic segmentation and holistic 3D reconstruction for the generic indoor and outdoor scenes from a single RGB image.
    \item In addition, we have developed an end-to-end pipeline for panoptic 3D parsing, Panoptic3D (end-to-end), where datasets have complete segmentation and 3D reconstruction ground-truth annotations.
\end{itemize}
Observing the experiments, we show encouraging results for indoor  \cite{fu20203dfuture} and the outdoor scenes for the natural scenes \cite{Lin2014COCO,Cordts2016Cityscapes}.

\section{Related Work \label{sec:related}}
Table \ref{tab:method_comparison} shows a comparison with related work. Our panoptic 3D parsing framework has the most complete set of properties and is more general than the competing methods. Next, we discuss related work below in details.

\noindent{\bf Single-view 3D scene reconstruction}.
Single image 3D reconstruction has a long history \cite{roberts1963machine,han2004automatic,Tulsiani17factored3d,zou2018layoutnet,huang2018cooperative,nie2020total3d}. The work in\cite{huang2018cooperative} jointly predicts 3D layout bounding box, 3D object bounding box, and camera intrinsics without any geometric reconstruction for indoor scenes. Factored3D \cite{Tulsiani17factored3d} is closely related to our work, which combines indoor scene layout (amodal depth) with 3D instance reconstructions without much abstraction. Still, no label is predicted for the scene layout (``stuff'') \cite{Tulsiani17factored3d}, and the instance object reconstruction tends to overfit the canonical shape of known categories. Total3DUnderstanding \cite{nie2020total3d} infers a box layout and has produced 3D reconstruction inference results on natural indoor images. However, as discussed before, these methods do not perform holistic 3D reconstruction for natural outdoor scenes or panoptic segmentation in general.

\noindent{\bf Single image depth estimation}. 
David Marr pioneered the 2.5D depth representation \cite{marr1982vision}. Depth estimation from a single image can be performed in a supervised way and has been extensively studied in the literature \cite{Saxena2009make3d,Eigen2014depth}. 
Development in deep learning \cite{long2015fully} has expedited the progress for depth estimation \cite{Bansal2016normal,Li2018megadepth}.
In our work, we adopt a relatively lightweight inverse depth prediction module from Factored3D \cite{Tulsiani17factored3d} and regress the loss jointly with 3D reconstruction and panoptic segmentation.

\noindent {\bf Single-view single object reconstruction}.
Single image single object 3D reconstruction can typically be divided into volume-based \cite{wu2015shapenets,zhang2018genre,chen2020topology}, mesh-based \cite{wang2018pixel2mesh,groueix2018atlas,kanazawa2018end}, and implicit-function-based \cite{chen2019learning, xu2019disn} methods.
In this paper, we adopt the detection and shape reconstruction branch from Mesh R-CNN \cite{gkioxari2019meshrcnn} for multi-object prediction . Building on top of it, we can perform supervised end-to-end single image panoptic 3D parsing. We also adopt unseen class reconstruction, GenRe \cite{zhang2018genre}, for multi-object reconstruction for natural image reconstruction when well-aligned ground truth 3D mesh models are not available.

\noindent{\bf Panoptic and instance segmentation}. Panoptic segmentation \cite{Kirillov2018panoptic} or image parsing \cite{tu2005image} combines semantic segmentation and instance detection/segmentation.
In our work, we build our panoptic head by referencing the end-to-end structure of UPSNet \cite{xiong19upsnet}. Additionally, we predict the 3D reconstruction of instances for each corresponding instance mask. However, the instance segmentation in panoptic segmentation is occluded. In comparison,
amodal instance segmentation predicts un-occluded instance masks for ``things''.
In this work, we generate both amodal as well as panoptic segmentation annotations from the 3D-FRONT dataset. This dataset enables the network to jointly perform 3D ``things'' reconstruction as well as panoptic segmentation. In the stage-wise pipeline, we utilize the work from Zhan \etal. \cite{zhan2020self} to better assist 3D reconstruction on natural images.

%%%%%%%%% METHOD
\section{Method}
We design our networks with the following goals in mind: 1). The network should be generalizable to both indoor and outdoor environments; 2). Datasets with various levels of annotations should be able to utilize the framework with simple replacement; 3). The segmentation masks should align with the reconstruction from the input view.

We will first introduce our stage-wise system, Panoptic3D (stage-wise) and show that it can process natural images without corresponding 3D annotations in training. Then, we will present our end-to-end network, Panoptic3D (end-to-end).

\subsection{Stage-wise Panoptic3D}

\begin{figure}[!ht]
\vspace{-2mm}
    \centering
    \includegraphics[width=\linewidth]{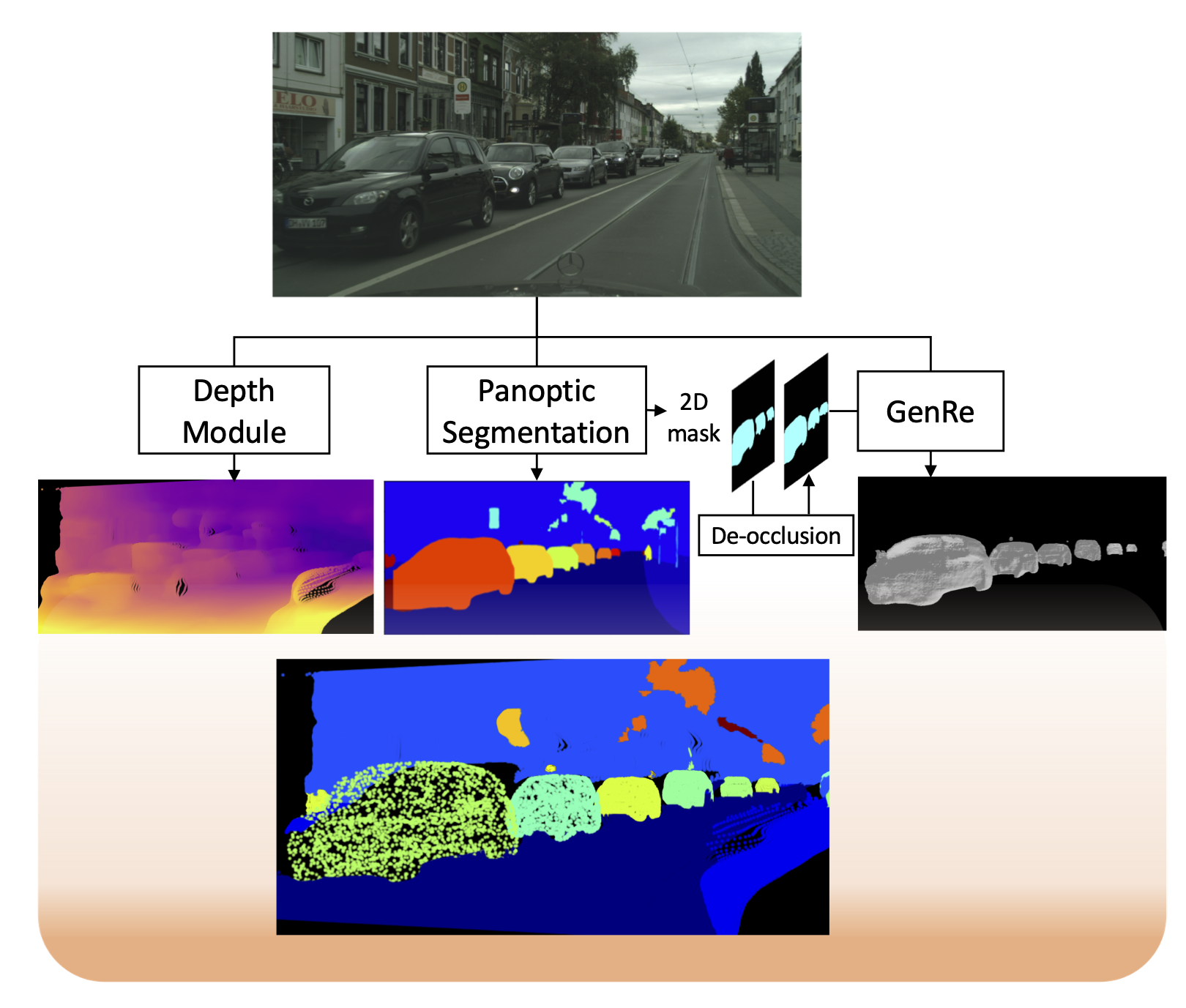}
    \caption{Our stage-wise system, Panoptic3D (stage-wise). We adopt DenseDepth \cite{Alhashim2018densedepth} for depth prediction, UPSNet \cite{xiong19upsnet} for panoptic segmentation, a de-occlusion network \cite{zhan2020self} for amodal mask completion, and GenRe \cite{zhang2018genre} to perform instance-based single image 3D reconstruction. The alignment module in Panoptic3D (stage-wise) outputs the image on the bottom.}
    \label{fig:networkstructure2}
\end{figure}

We present our stage-wise system, Panoptic3D (stage-wise), in Figure~\ref{fig:networkstructure2}. We design this system for natural image datasets that contain well-annotated panoptic segmentation information but lack 3D information, such as COCO \cite{Lin2014COCO} and Cityscapes \cite{Cordts2016Cityscapes}. This stage-wise system contains four main parts: 1). Instance and panoptic segmentation network. 2). Instance amodal completion network. 3). Single object 3D reconstruction network for unseen classes. 4). Single-image depth prediction network.

We take advantage of the state-of-art panoptic segmentation system, UPSNet \cite{xiong19upsnet}, scene de-occlusion system \cite{zhan2020self}, depth prediction system, DenseDepth \cite{Alhashim2018densedepth}, and unseen class object reconstruction networks \cite{zhang2018genre} and integrate them into a single pipeline.

This Panoptic3D (stage-wise) framework takes an RGB image and predicts the panoptic 3D parsing of the scene in point cloud (for ``stuff'') and meshes (for `` things''). The implementation details are as follows: the network first takes panoptic results from UPSNet and depth estimation from DenseDepth. It then passes the modal masks to the de-occlusion net to acquire amodal masks. We use GenRe to reconstruct the instance meshes based on the amodal masks. The module then maps panoptic labels to depth pixels and uses empirically estimated camera intrinsics estimation to inverse project depth into point clouds. 

Since GenRe \cite{zhang2018genre} only predicts normalized meshes centering at the origin, the final module aligns individual shapes using depth estimation in the z-direction and the mask in the x-y direction. The module takes the mean of the $98th$ percentile and the $2nd$ percentile of the filtered and sorted per-pixel depth prediction within the predicted mask region to estimate the z center depth of each object. Finally, it places meshes and depth point cloud in the same coordinate system to render the panoptic 3D parsing results. The general inference time is 2.4 seconds per image on one NVIDIA Titan X GPU.

\subsection{End-to-end Panoptic3D}

\begin{figure*}[!ht]
    \centering
    \includegraphics[width=0.9\textwidth]{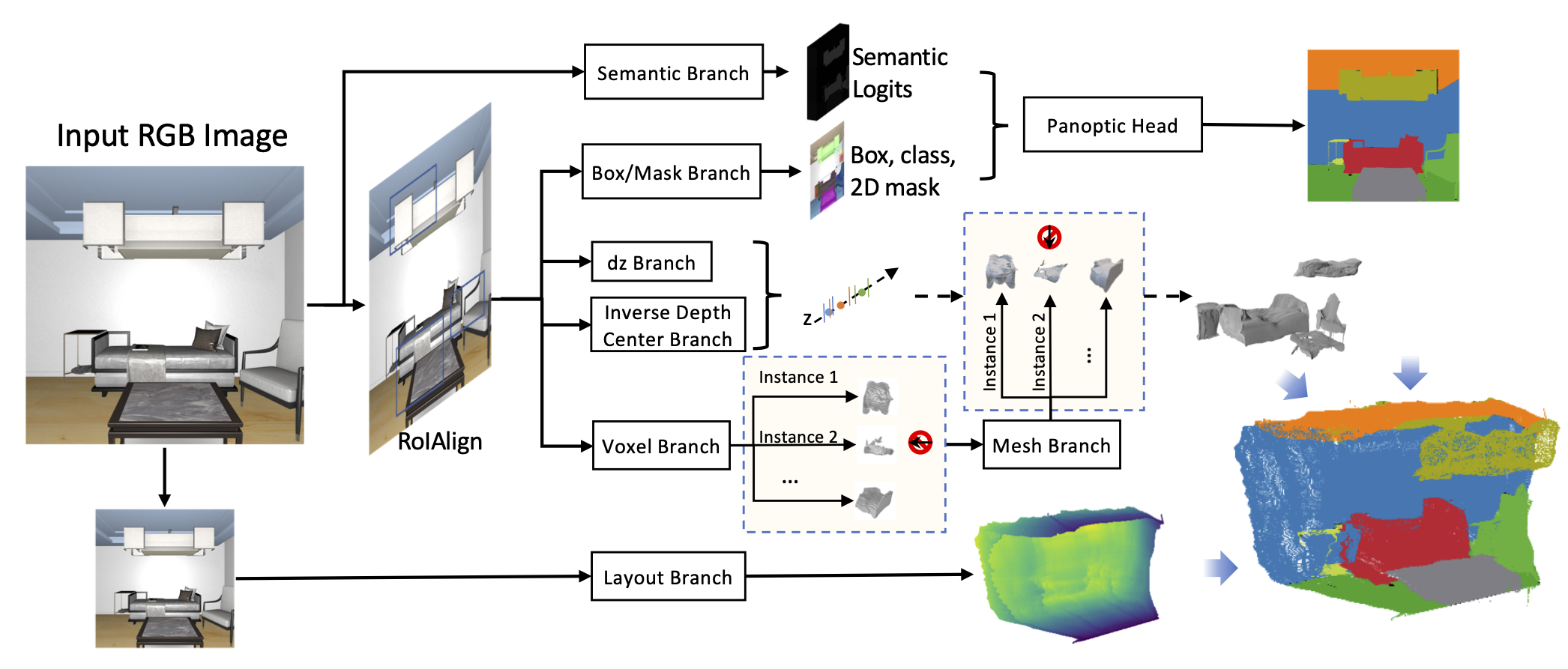}
    \caption{Network architecture for our Panoptic3D (end-to-end) pipeline. dz: means depth extent. The red stop sign indicates that only predictions with centered ground truth shapes are used for regression during training. }
    \label{fig:networkstructure1}
    \vspace{-2mm}
\end{figure*}

The overview of the end-to-end network structure, Panoptic3D (end-to-end), is shown in Figure~\ref{fig:networkstructure1}. Similar to Panoptic3D (stage-wise), Panoptic3D (end-to-end) also has four main components: 1). instance segmentation head. 2). multi-object training enabled shape heads. 3). panoptic segmentation head. 4). ``stuff'' depth and relative object z center prediction branch. The entire network is trained end-to-end and can jointly predict amodal instance, semantic, and panoptic segmentation, ``stuff'' depth, and ``things'' reconstruction in 3D. Our design ideas are as follows.

For the panoptic 3D prediction, we predict ``stuff'' depth instead of box representation because it is not easily generalizable to scenes with other ``stuff'' categories, such as windows, doors, rugs, etc., and it does not apply to outdoor environments. Taking advantage of the advanced development in 2D panoptic segmentation, we first predict 2D panoptic segmentation and then align ``stuff'' segmentation with the depth prediction.

We predict that amodal ``stuff'' prediction would significantly improve the panoptic 3D parsing task for future works.

For multi-object 3D reconstruction, we first enable multi-object training and evaluation for the baseline network. For joint training with panoptic segmentation, we mainly address the following three challenges: 1). With multiple objects in a scene, mesh shapes that are too close to the camera may have a negative z-center, not tolerated in end-to-end detection and reconstruction baseline model by design. 2). For objects that appear to be cut-off by the camera view (non-centered/boundary objects) or too close to the camera, transformation to camera coordinate will deform ground truth 3D voxel and mesh into a shape that contain infinitely far points, preventing the network from converging. One approach would be to cut the ground truth shapes to be within the camera frustum. However, this may result in unnatural edge connections, and the preprocessing step is time-consuming. Instead, we introduce a partial loss. We first detect and mark objects occluded by image boundaries (non-centered/boundary objects) and exclude their loss for shape-related regressions. For example, we use an indicator function $\mathds{1}(\cdot)$ to return 1 for centered objects and 0 for boundary objects. The final loss per batch is defined as $\mathcal{L} = \mathcal{L}_{mask}$\cite{he2017mask} $+ \mathcal{L}_{box}$ \cite{he2017mask} $+ \mathcal{L}_{class}$ \cite{he2017mask} $+ \mathcal{L}_{panoptic}$ \cite{xiong19upsnet}$ + \mathcal{L}_{semantic}$ \cite{Kirillov2019fpn} $+ \mathcal{L}_{depth}$ \cite{Tulsiani17factored3d} $+ \mathds{1} \cdot (\mathcal{L}_{dz}$ \cite{gkioxari2019meshrcnn} $+ \mathcal{L}_{zc} + \mathcal{L}_{voxel}$ \cite{gkioxari2019meshrcnn} $+ \mathcal{L}_{mesh}$ \cite{gkioxari2019meshrcnn}$)$.

For depth, we use a simple U-Net network structure to predict inverse ``stuff'' depth, which is adopted from Factored3D \cite{Tulsiani17factored3d} because it is relatively lightweight. In addition to depth, to assist the positioning of objects relative to their environment, we add an inverse z center prediction head to align predicted objects with the predicted layout or depth map in 3D. The z center is defined as $z_c$ in $\bar{dz} = \frac{d_z}{z_c} \cdot \frac{f}{h}$, where $\bar{dz}$ is defined as the scale-normalized depth extent \cite{gkioxari2019meshrcnn}, $h$ is the height of the object's bounding box, $f$ is the focal length, $d_z$ is the depth extent. Our z center head predicts the inverse $z_c$, which is the object's center in the z-axis of the camera coordinate system.

In summary, the Panoptic3D (end-to-end) network uses a ResNet and an FPN network as our backbone for detection, along with an FPN-based semantic head to assist the 2D panoptic prediction, an inverse z center head in predicting object centers relative to inverse depth prediction produced by the depth branch, and enables multi-object training and evaluation for the shape heads.

\begin{table*}[!ht]
    \centering
        \caption{Available datasets comparison. More comparison is available in \cite{fu20203dfuture}. The last row shows the panoptic 3D 3D-FRONT dataset rendered and annotated by us.}
    \label{tab:inhouse_dataset}
    \scalebox{0.9}{
    \begin{tabular}[width=\linewidth]{cccccccc}
    \hline
         Dataset & Instance & Semantic & Panoptic &Depth & 3D ``things'' & 3D ``stuff'' & Alignment\\
         \hline
         SUN-RGBD\cite{song2014sunrgbd} &\cmark&\cmark&-&\cmark&0&-&-\\
         AI2Thor\cite{Kolve2017AI2THORAn}&\cmark&\cmark&\cmark&\cmark&100&\cmark&\cmark\\
         ScanNet\cite{dai2017scannet}&\cmark&\cmark&-&\cmark&14225/1160\cite{Avetisyan2019scan2cad}&-&approx.\cite{Avetisyan2019scan2cad}\\
         3D-FUTURE\cite{fu20203dfuture}&\cmark&-&-&-&9992&-&\cmark\\
         3D-FRONT\cite{fu20203dfuture}&-&-&-&-&9992&\cmark&\cmark\\
         Panoptic 3D 3D-FRONT&\cmark&\cmark&\cmark&\cmark&2717&\cmark&\cmark\\
         \hline
    \end{tabular}
    }
\end{table*}

\begin{figure*}[!ht]
    \centering
    \includegraphics[width=0.9\linewidth]{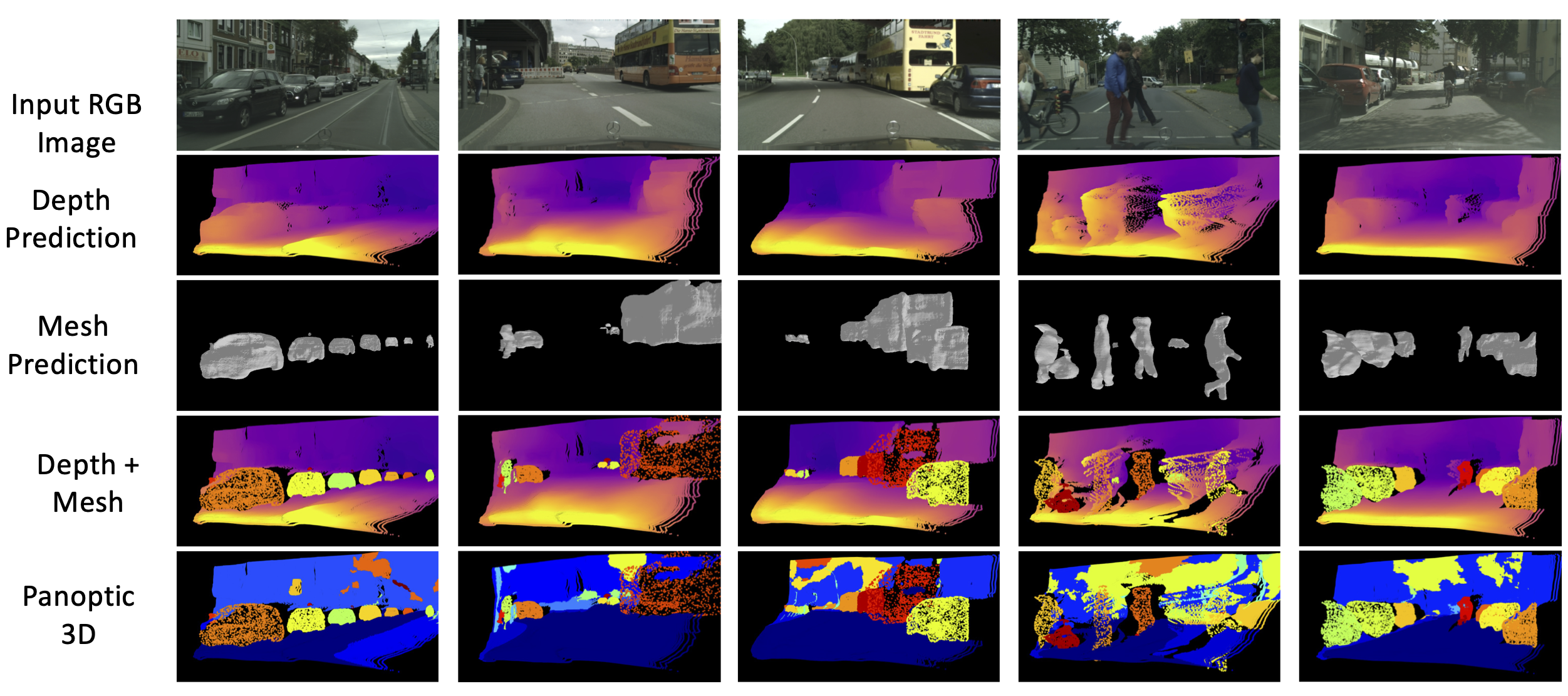}
    \caption{Qualitative results of Panoptic3D (stage-wise) on Cityscapes images \cite{Cordts2016Cityscapes}. Results are taken from an off-angle shot to show the difference between depth and 3D panoptic results. We sampled the point cloud from the result object meshes for better visualization of the 3D effect. We show that our alignment module outputs visually correct alignment between things reconstruction and ``stuff'' depth point cloud.}
    \label{fig:cityscapes}
    \vspace{-4mm}
\end{figure*}

\section{Datasets}
For Panoptic3D (stage-wise), we show qualitative results for natural datasets such as COCO and Cityscapes, where well-annotated panoptic segmentation labels are provided. 

To our best knowledge, no available dataset is accurately annotated with amodal instance segmentation, panoptic segmentation, 2.5D information for ``stuff'', and 3D meshes for ``things''. 
Most natural image datasets either do not provide panoptic segmentation annotations or suffer from low diversity or low quantity for corresponding 3D mesh annotations. ScanNet \cite{dai2017scannet} has a diverse environment, a large number of images annotated with instance/semantic segmentation, and annotations for corresponding 3D meshes. However, the mesh annotations on ScanNet do not have good alignment with their masks. Additionally, our attempt to generate panoptic segmentation information for ScanNet suffers from significant human errors in semantic and instance segmentation annotations. Therefore, we are not able to work on ScanNet for the current end-to-end supervised system. We are also aware of other 3D datasets such as SUN-RGBD \cite{song2014sunrgbd}, AI2Thor \cite{Kolve2017AI2THORAn}, Scan2CAD \cite{Avetisyan2019scan2cad}, 3D-FUTURE \cite{fu20203dfuture} and OpenRooms \cite{Li2020OpenRoomsAE}. We show in Table~\ref{tab:inhouse_dataset} that the natural datasets, such as SUN-RGBD and ScanNet, do not precisely align 3D ``stuff'' or ``things''. For a virtual dataset, even though we can extract all the information from AI2Thor, the number of shapes was too limited for shape reconstruction training during the early stages of our project. OpenRooms has not yet released its 3D CAD models.

Thanks to the availability of the 3D-FRONT dataset \cite{fu20203dfuture}, we can generate a first version of the panoptic 3D parsing dataset with COCO-style annotations, including 2D amodal instance and panoptic segmentation, modal and amodal (layout) depth, and corresponding 3D mesh information for every image. Referenced from the 3D-FUTURE dataset \cite{fu20203dfuture}, we adopt 34 instance categories representing all of the countable objects as ``things'', and add three categories representing walls, ceilings, and floors as ``stuff'' as no other ``stuff'' categories exist in the first release.

For rendering, the first release of the 3D-FRONT dataset does not provide the textures and colors for ``stuff'' objects, so we adopt textures from the SceneNet RGB-D dataset \cite{McCormac2017scenenet}. We place a point light at the renderer's camera position for the lightning to make sure the scene is fully lit. We use the official Blender\cite{blender} script with the officially released camera angles for this work.

\begin{figure*}[!ht]
    \centering
    \includegraphics[width=0.7\linewidth]{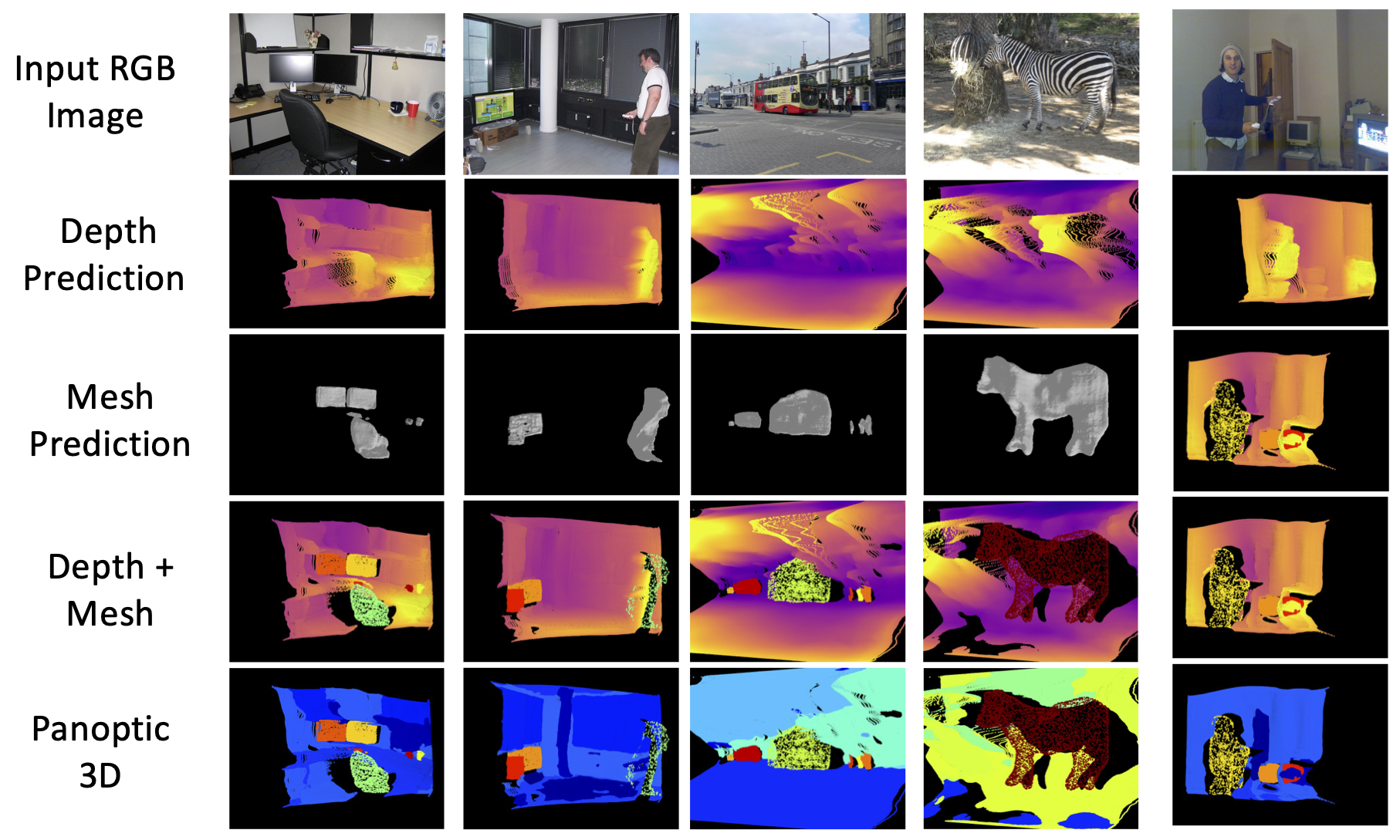}
    \caption{Qualitative results for Panoptic3D (stage-wise) on COCO images \cite{Lin2014COCO}. Results are taken from an off-angle shot to show the difference between depth and 3D panoptic results. We sampled the point cloud from the predicted object meshes for better visualization of 3D structures. }
    \label{fig:COCO}
    \vspace{-4mm}
\end{figure*}

We use the first 1620 houses as the train set and the last 200 houses as the test set for the experiments. We first mark all objects that appear both in the train and test sets as invalid during training, ensuring that the 3D models are disjoint between the train and test sets. We only train and evaluate our mesh prediction on non-boundary (or relatively centered) objects. After filtering out images with no valid things, there are 7734 images in the train set and 1086 in the test set. There are 55216 instances in the train set for the panoptic segmentation task and 7548 instances in the test set. For the 3D reconstruction task, there are 1559 unique models in the train set and 1158 unique models in the test set. The final dataset covers 33 categories for ``things'' during training and 31 types of ``things'' during evaluation.

\section{Experiment Details and Evaluation}

\begin{table*}[!ht]
\centering
\caption{\small \textbf{Comparison of re-projected 2D panoptic qualities from a subset of COCO indoor images between Total3DUnderstanding and our Panoptic3D (stage-wise) system.} For Total3DUnderstanding, the re-projection uses inferred camera extrinsic and we change the predicted layout box into meshes for wall, ceiling, and floor. Our Panoptic3D (stage-wise) method outperforms Total3DUnderstanding on every metrics.}
\label{tab:total3dpanoptic}
\vspace{-2mm}
    \scalebox{0.85}{
    \begin{tabular}{c|ccc|ccc|ccc}
\hline
  & \multicolumn{3}{c}{PQ $\uparrow$}  & \multicolumn{3}{c}{SQ $\uparrow$} & \multicolumn{3}{c}{RQ $\uparrow$} \\
  
Methods & IOU@.5 & IOU@.4 & IOU@.3  & IOU@.5  & IOU@.4  & IOU@.3  & IOU@.5 & IOU@.4  & IOU@.3 \\ 
\hline
Total3DUnderstanding \cite{nie2020total3d}   & 0.043 & 0.06 & 0.077 & 0.046 & 0.063 & 0.081 & 0.065 & 0.101 & 0.15\\
Panoptic3D (stage-wise) (ours)   & \textbf{0.168} & \textbf{0.176} & \textbf{0.181} & \textbf{0.177} & \textbf{0.184} & \textbf{0.181} & \textbf{0.21} & \textbf{0.220} & \textbf{0.226}\\ 
\bottomrule
\end{tabular}
}
\end{table*}

\subsection{Stage-wise Panoptic3D}

Datasets such as COCO and Cityscapes, have well-annotated panoptic segmentation annotations but lack annotations of 3D shapes and depth information. Figure~\ref{fig:networkstructure2} shows the stage-wise system pipeline. With UPSNet\cite{xiong19upsnet} as the backbone, we can use a de-occlusion network \cite{zhan2020self} for amodal mask prediction and a depth network \cite{Alhashim2018densedepth} and an alignment module for scene alignment. Additionally, we use the predicted amodal mask and the input RGB image for unseen class instance reconstruction \cite{zhang2018genre}. The outputs of these networks would then be passed through an alignment module that produces the 3D panoptic parsing results.

For the Cityscapes dataset, we compute its camera intrinsics with FOV = 60, height = 1024 and width = 2048 \cite{Cordts2016Cityscapes}. Since it doesn't provide camera information for the COCO dataset, we estimate its FOV to be 60 based on heuristics and use an image size of $480 \times 640$, which is compatible with every sub-module of the stage-wise system. 

In Figure~\ref{fig:cityscapes} and Figure~\ref{fig:COCO}, we show qualitative measures for Cityscapes and COCO, respectively. The pipeline has demonstrated qualitatively good results for both indoor and outdoor natural images.

Using the COCO dataset, we can project the panoptic 3D results back to the input view and evaluate it against their ground truth 2D panoptic annotation to show its image parsing capability. We acquired around 300 images from the COCO test set that contains overlapped panoptic labels Total3DUnderstanding. In Table~\ref{tab:total3dpanoptic}, we show that our pipeline outperforms Total3DUnderstanding on reprojected panoptic segmentation metrics.

\subsection{End-to-end Panoptic3D}
We train our networks with a learning rate of 0.005 for 30000 iterations. We use PyTorch for code development and 4 GeForce GTX TITAN X GPUs for ablation studies. We switch to 8 GPUs for larger architectures with depth/layout predictions. The experiments with the largest model take 16 hours to run on 8 GeForce GTX TITAN X GPUs. Our input size for the detection backbone is $1024 \times 1024$ instead of the original $800 \times 800$ used by Mesh R-CNN because the depth network requires the input to be divisible by 64. The input image is resized to $512 \times 512$ for the depth branch. The final network contains 13 losses: semantic segmentation pixel-wise classification loss, panoptic segmentation loss, RPN box classification loss, RPN box regression loss, instance box classification loss, instance box regression and segmentation loss, depth extent loss, inverse depth center loss, voxel loss, mesh loss, depth loss, and ``stuff'' depth loss. Partial loss is used for depth extent, object inverse depth loss, voxel loss, and mesh loss.

\subsection*{Shape Reconstruction}
\begin{figure*}
    \centering
    \includegraphics[width=0.8\linewidth]{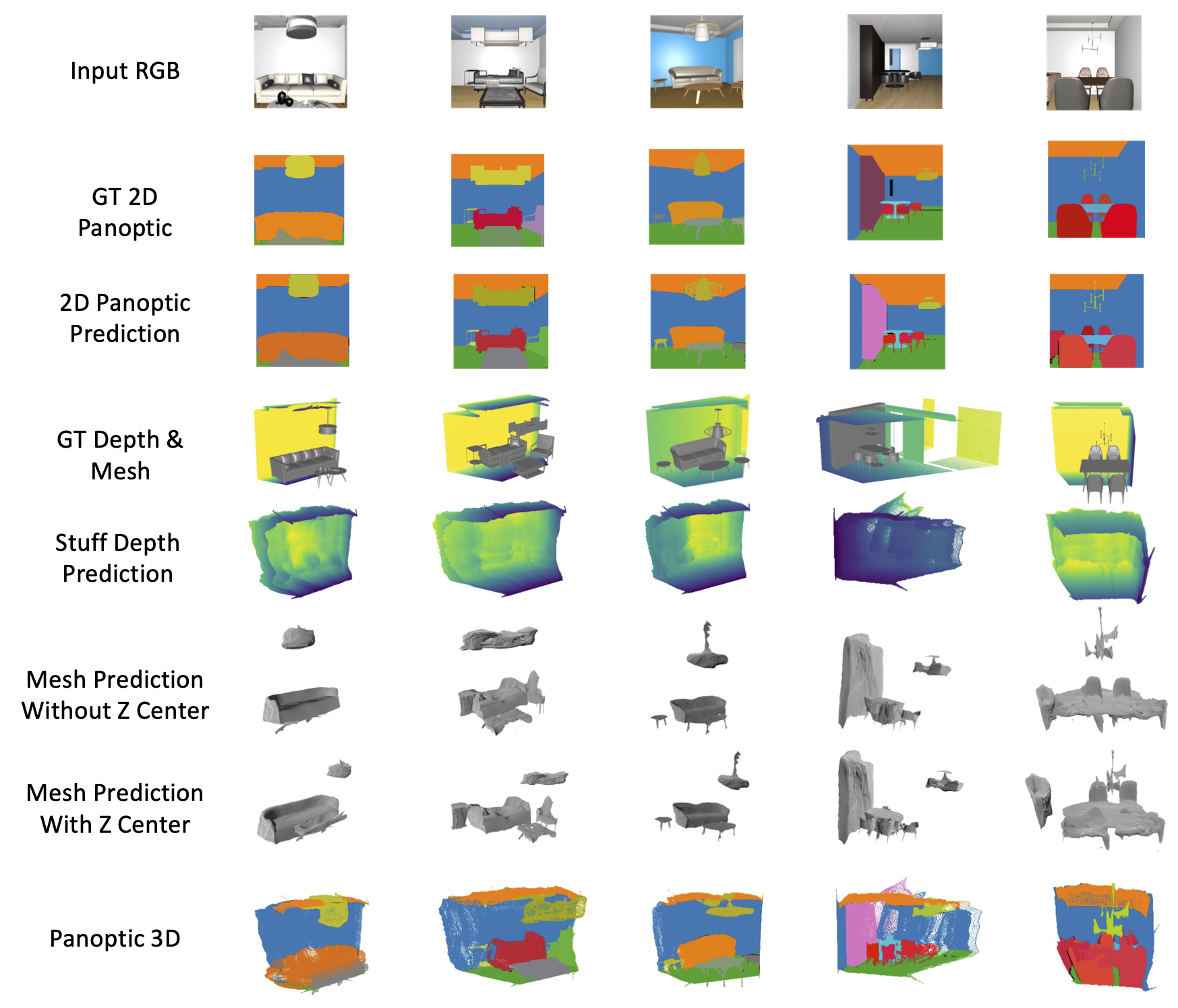}
    \caption{Qualitative results for Panoptic3D (end-to-end) on the 3D-FRONT dataset \cite{fu20203dfuture}. We show our predicted panoptic results (row 3) compared to panoptic ground truth (row 2) and our reconstruction results (row 5 to 7) to reconstruction ground truth (row 4). Row 8 shows the final panoptic 3D results, where we sample point clouds from meshes for better visualization. Comparing rows 6 and 7, row 7 shows the object placement is more consistent with the input RGB image. Our final results show that the shape predictions align well with the predicted ``stuff'' depth prediction.}
    \label{fig:3dfront}
\end{figure*}

For the baseline model, we add multi-instance training and allow shape regression only on centered objects on top of detection and reconstruction network structures used in \cite{gkioxari2019meshrcnn}.
Ablations on partial-loss training and joint training with other heads are included in Table~\ref{tab:baselines} and Table~\ref{tab:shape}. We find that utilizing more samples per image for training the instance head can help improve mesh prediction with higher  $AP^{mesh}$ in Table~\ref{tab:baselines}. In Table~\ref{tab:shape}, we show that adding additional panoptic, z-center, depth, and layout heads significantly improve the average precision for boxes and masks, but only a slight improvement on meshes when used together. Notice that adding z-center loss starting from the model (b) does not significantly boost the earlier models; however, it provides considerably better qualitative visualization in Figure~\ref{fig:3dfront}. Compared to row 6 (without z-center loss), row 7 (with z-center loss) shows a more consistent layout against the input RGB image. The furniture cluster around a similar depth in row 6.

\begin{table}[!ht]
    \centering
    \caption{Baseline model comparisons. Our baseline model (1) is a multi-object training and evaluation enabled detection and reconstruction network \cite{gkioxari2019meshrcnn} trained on on centered objects in all three heads (instance, voxel, mesh). Model (2) is the baseline model trained on all things with a partial loss on voxel and mesh heads. N indicates the number of annotations used to regress each corresponding head during training. }
    \label{tab:baselines}
    \scalebox{0.65}{
    \begin{tabular}{c|ccc|ccc}
        \hline
         Baseline&N&N&N&$AP^{box}$ & $AP^{mask}$ & $AP^{mesh}$  \\
         &instances&voxels&meshes&&&\\
         \hline
         (1)&16175&16175&16175& 37.8 $\pm$ 1.4& 34.2 $\pm$ 1.9& 5.9 $\pm$ 0.4\\
         (2)&55216&16175&16175& 56.5 $\pm$ 0.9& 52.6 $\pm$ 1.1& 8.9 $\pm$ 1.5\\
         \hline
    \end{tabular}
    }
\end{table}

\begin{table}[ht]
    \centering
    \caption{Ablation studies for the Panoptic3D (end-to-end) model on the panoptic 3D 3D-FRONT dataset. Our ablation study is compared with the baseline (2) in Table~\ref{tab:baselines}. The results show that during joint training, the network can maintain its $AP^{mesh}$ performance while improving on $AP^{box}$ and $AP^{mask}$.}
    \label{tab:shape}
        \scalebox{0.66}{
    \begin{tabular}{c|cccc|ccc}
    \hline
        &panoptic &z-center & depth& layout &$AP^{box}$ & $AP^{mask}$ & $AP^{mesh}$\\
            \hline
        (2)& -&-&-&- & 56.5 $\pm$ 0.9& 52.6 $\pm$ 1.1& 8.9 $\pm$ 1.5 \\
    \hline
        (a) & \cmark&&& &56.7 $\pm$ 2.2& 55.8 $\pm$ 2.7 & 8.3 $\pm$ 0.6 \\
        (b) & \cmark&\cmark&&& 57.0 $\pm$ 1.7& 55.5 $\pm$ 1.2  & 8.1 $\pm$ 1.5\\
        (c) & \cmark&\cmark&\cmark& &59.4 $\pm$ 0.6&56.7 $\pm$ 2.4 & 9.0 $\pm$ 1.2\\
     \hline
       ours &\cmark &\cmark&\cmark&\cmark&\textbf{60.0 $\pm$ 1.4}&\textbf{56.0 $\pm$ 1.4}&\textbf{9.0 $\pm$ 1.3} \\
        \hline
    \end{tabular}
}
\end{table}

\subsection*{Panoptic Segmentation}
We compare our panoptic segmentation results with the original UPSNet panoptic segmentation results as one of our baselines. Although we use a panoptic feature pyramid network \cite{Kirillov2019fpn} instead of the FPN network with deformable CNNs from UPSNet, the results are comparable. We notice a slight decrease when we switch masks for instance head from modal to amodal, as amodal masks may pose challenges to the panoptic head. As for joint training, we show in Table~\ref{tab:panoptic} that the results from joint training are comparable with our baseline. We use the metrics of PQ, SQ, and RQ following the panoptic segmentation paper \ref{tab:panoptic}.

We are aware that our ``stuff'' categories are an easy set for panoptic segmentation tasks. 3D-FRONT offers new releases from when we began the project, so for future studies, we will attempt to incorporate more categories, such as doors and windows, with better rendering effects. However, our dataset does provide the first version of any such dataset that enables end-to-end training on the task of panoptic 3D parsing. Based on our acquired results, there are still challenges even with the current ``stuff'' categories for panoptic segmentation.

\begin{table}[ht]
    \centering
    \caption{Ablation studies for the Panoptic3D (end-to-end) model on the panoptic 3D 3D-FRONT dataset. The model numbers here correspond to models in Table~\ref{tab:shape}. Here we show that the panoptic performance is comparable to the baseline performance with joint training.}
    \label{tab:panoptic}
    \scalebox{0.8}{
\begin{tabular}{c|cccc|ccc}
    \hline
        &panoptic &z-center & depth& layout &PQ & SQ & RQ\\
        \hline
        (a) & \cmark&&& &46.4&76.8&54.0 \\
        (b) & \cmark&\cmark&&& 46.0& 75.9 &53.4 \\
        (c) & \cmark&\cmark&\cmark& &47.4&76.1 & 55.2\\
     \hline
       ours &\cmark &\cmark&\cmark&\cmark&46.9&75.7&54.4 \\
        \hline
    \end{tabular}
    }
\end{table}

\subsection*{Depth and Layout predictions}
We adopt the U-Net structure from Factored3D and jointly train the depth with the rest of our pipeline. We find regressing for layout depth alone is an easier task for the network than joint training. Joint training using layout depth loss with other losses appear to be a challenging problem. Adding cross-consistency loss \cite{zamir2020consistency} between layout depth and normal does not seem to improve depth's performance easily. Nonetheless, adding layout depth loss appears to help the network to perform better in general in both shape metrics and panoptic metrics as shown in Table~\ref{tab:shape} and Table~\ref{tab:panoptic}.

\vspace{-2mm}
\section{Conclusions}
\vspace{-2mm}

This paper presents a framework that aims towards tackling panoptic 3D parsing for a single image in the wild. We demonstrate qualitative results for natural images from datasets such as Cityscapes \cite{Cordts2016Cityscapes} (shown in Figure~\ref{fig:cityscapes}) and COCO \cite{Lin2014COCO} (shown in Figure~\ref{fig:COCO}) under the stage-wise system. Additionally, an end-to-end pipeline that can be trained with full annotations is also proposed.

For societal impact, we are proposing a new task in computer vision, Panoptic 3D Parsing (Panoptic3D), that is concerned with a central task in computer vision: joint single image foreground object detection and segmentation, background semantic labeling, depth estimation, 3D reconstruction, and 3D layout estimation. A successful Panoptic3D system can see its applications in a wide range of domains beyond computer vision such as mapping, transportation, computer graphics etc. However, Panoptic3D is a learning based system and may have bias introduced in various training stages. Careful justification and adoption of the system in appropriate tasks subject to regulation are needed.\\

\noindent {\bf Limitations}: We use a fixed FOV of 60 and fixed image sizes for Cityscapes, COCO, and 3D-FRONT. Hence the estimated 3D scene can only provide a rough ordering estimation of things and stuff in terms of distance to the camera plane. We use a single light source for the 3D-FRONT image rendering, which results in artificial lighting for the training images. Therefore, the generalization capability for the models trained using this dataset might not be strong. We expect the model to generalize better under more natural lighting conditions, which is to be verified in the next step with a new dataset.

\section{Acknowledgments}
Part of the work was done during Sainan Liu's internship at Intel.

%%%%%%%%% REFERENCES
{\small
\bibliographystyle{ieee_fullname}
\bibliography{main}
}

\end{document}